%% file: arxiv.tex
\def\isarxivversion{1}

\ifdefined\isarxivversion\else
\def\isreviewversion{1}
\fi
\documentclass[10pt,twocolumn,letterpaper]{article}

\ifdefined\isarxivversion
\usepackage[pagenumbers]{cvpr}
\else
\usepackage[review]{cvpr}      
\fi

\input{preamble}
\ifdefined\isarxivversion
\usepackage{microtype}
\makeatletter
\def\cvprsection{\@startsection{section}{1}{\z@}
   {-9pt plus -2pt minus -2pt}{6pt}{\large\bf}}
\def\cvprsubsection{\@startsection{subsection}{2}{\z@}
   {-7pt plus -2pt minus -2pt}{4pt}{\elvbf}}
\makeatother
\fi
\ifdefined\issupplementaryversion
    \usepackage{xr}
    \makeatletter
    \begingroup
    \let\bibcite\@gobbletwo
    \endgroup
    \makeatother
\fi

\definecolor{cvprblue}{rgb}{0.21,0.49,0.74}
\usepackage[pagebackref,breaklinks,colorlinks,allcolors=cvprblue,hypertexnames=false]{hyperref}

\def\confName{CVPR}
\def\confYear{2027}
\title{Visual Token Coding for Video Multimodal Large Language Models}
\ifdefined\isarxivversion
\input{authors}
\else
\author{Anonymous Submission}
\fi
\hypersetup{
    pdftitle={Visual Token Coding for Video Multimodal Large Language Models},
    pdfauthor={}
}
\ifdefined\isarxivversion
\hypersetup{pdfauthor={Chenxin Fang, Tao Chen, JunChao You, Jun Peng, Yiyi Zhou, Rongrong Ji}}
\fi

\begin{document}
\raggedbottom
\setlength{\dbltextfloatsep}{10pt plus 2pt minus 2pt}
\ifdefined\isarxivversion
\setlength{\textfloatsep}{14pt plus 2pt minus 2pt}
\setlength{\floatsep}{10pt plus 2pt minus 2pt}
\setlength{\dbltextfloatsep}{10pt plus 2pt minus 2pt}
\setlength{\dblfloatsep}{10pt plus 2pt minus 2pt}
\fi
\ifdefined\issupplementaryversion
\setcounter{figure}{5}
\setcounter{table}{3}
\input{sec/X_suppl}
\else
\begingroup
\makeatletter
\renewcommand{\@fnsymbol}[1]{\ensuremath{\dagger}}
\makeatother
\maketitle
\endgroup
\setcounter{footnote}{0}

\input{sec/0_abstract}
\input{sec/1_intro}

\input{sec/2_related_work}
\input{sec/3_preliminary}
\input{sec/4_method}
\input{sec/5_experiments}
\input{sec/6_conclusion}

\FloatBarrier
\ifdefined\isarxivversion
\pagebreak
\else
\clearpage
\fi
\raggedbottom
\input{sec/llm_usage}
\fi
\FloatBarrier
{\small
    \bibliographystyle{ieeenat_fullname}
    \bibliography{main}
}

\end{document}

%% file: preamble.tex
\usepackage{multirow}
\usepackage{array}
\usepackage{colortbl}
\usepackage{placeins}
\ifdefined\issupplementaryversion
\usepackage{float}
\fi
\usepackage{cuted}
\usepackage{currfile}
\usepackage{caption}
\usepackage{xcolor}

\DeclareRobustCommand{\wordflag}[1]{#1}

\DeclareRobustCommand{\vtc}{%
  \texorpdfstring{\textmd{VTC}}{VTC}}
\DeclareRobustCommand{\vtcdy}{%
  \texorpdfstring{\textmd{VTC\textsubscript{dy}}}{VTC-Dy}}

\makeatletter
\renewcommand\LenToUnit[1]{\strip@pt\dimexpr#1*\p@/\unitlength\relax}
\makeatother

%% file: authors.tex
\author{
    Chenxin Fang, Tao Chen, JunChao You, Jun Peng,
    Yiyi Zhou\thanks{Corresponding author.}, Rongrong Ji\\
    Key Laboratory of Multimedia Trusted Perception and Efficient Computing,\\
    Ministry of Education of China, Xiamen University, 361005, P.R. China\\
    {\tt\small \{fangchenxin, chentao, youjunchao\}@stu.xmu.edu.cn}\\
    {\tt\small \{pengjun, zhouyiyi, rrji\}@xmu.edu.cn}
}

%% file: sec/0_abstract.tex
\begin{abstract}
In this paper, we propose a new token compression paradigm for video \emph{Multimodal Large Language Models} (MLLMs),
termed \emph{Visual Token Coding} (\textbf{VTC}).
Inspired by classical video coding principles, \emph{e.g.}, HEVC, \vtc{} performs structured compression by
predicting the I/P frames of a video and measuring their frame-wise residuals to estimate token redundancy.
Based on this base paradigm, we further extend \vtc{} with a set of dynamic budgeting designs towards more cost-effective token coding, termed \textbf{VTC\textsubscript{dy}}.
To validate \vtc{} and \vtcdy{}, we apply them to three MLLMs and conduct experiments on a set of video understanding benchmarks.
The experimental results show that
as a base paradigm, \vtc{} can achieve an average performance retention of 98.4\% on Qwen3-VL under a 50\% token compression ratio, and the enhanced \vtcdy{} can further improve the retention to 100.3\%, outperforming a set of compared methods.
Moreover, as a plug-and-play design, \vtc{} requires no additional tuning of MLLMs for token coding.
\ifdefined\isreviewversion
\textbf{Our code} is given in the supplementary materials.
\else
Our code is available at \url{https://github.com/Msr233/VTC}.
\fi
\end{abstract}

%% file: sec/1_intro.tex
\section{Introduction}
\begin{figure}[t]
    \centering
    \includegraphics[width=\columnwidth,keepaspectratio]{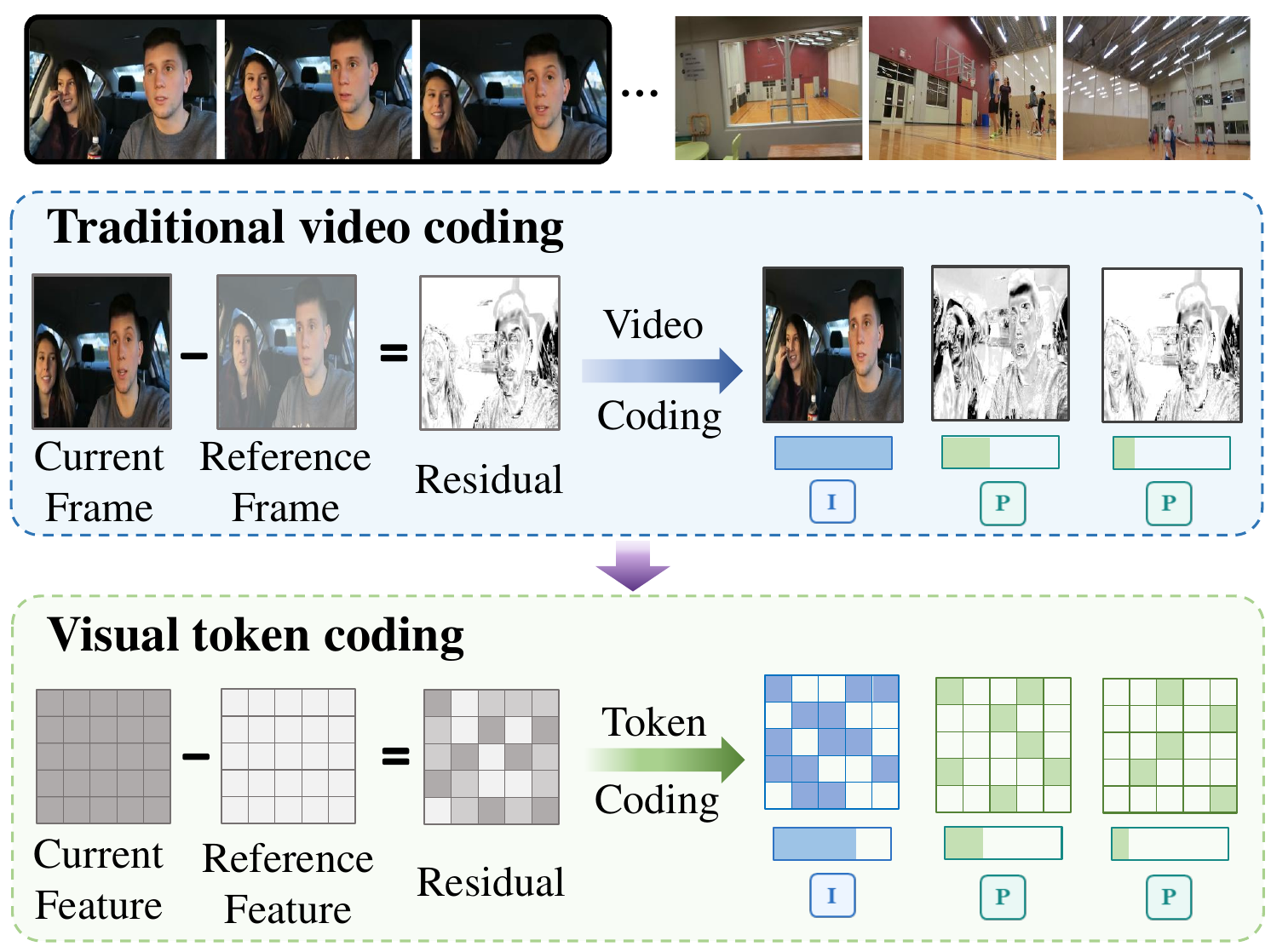}
    \caption{Motivation of \vtc{}. Traditional video coding predicts P frames from I-frame references and encodes residuals with different I/P-frame budgets. \vtc{} introduces this structured compression principle to the feature space of visual encoders and performs reference prediction and residual coding, which can be \wordflag{plugged into} existing MLLMs and requires no additional training.}
    \label{fig:motivation}
\end{figure}
In recent years, great breakthroughs have been made by advanced \emph{Multimodal Large Language Models} (MLLMs) in a variety
of video-language tasks \citep{maaz2024videochatgpt,jin2024chatunivi,lin2024videollava,cheng2024videollama2,li2024llavaonevision,zhang2024llavavideo,bai2025qwen25vl,bai2025qwen3vl}.
Despite their great success, existing video MLLMs are still hindered by the excessive number of visual tokens used to represent input videos \citep{tao2025dycoke,wu2026metom}, which greatly limits their practical applications. For instance, a cutting-edge MLLM such as Qwen3-VL \cite{bai2025qwen3vl} often requires up to $105k$ visual tokens to represent an hour-long video at 768 frames, yielding expensive computation and memory costs.

In this case, token compression has become a research hot spot for efficient MLLMs, \wordflag{whose paradigms include} \emph{token pruning} \cite{chen2024fastv,huang2025prunevid,alvar2025divprune,ye2025atpllava,yang2025topv}, \emph{token merging} \cite{bolya2023tome,yang2025visionzip,shao2025holitom,tao2025dycoke}, or their combinations \cite{fu2025framefusion,dhouib2025pact,yang2025libramerging,zhang2026hybrid,singh2026duet}.
In practice, most existing methods achieve token compression by measuring token-wise importance based on the metrics such as \emph{visual saliency} \cite{chen2024fastv}, \emph{token diversity} \cite{yang2025visionzip,alvar2025divprune}, or \emph{query relevance} \cite{huang2025prunevid,liu2025kvtp}. For video tasks, temporal relationship modeling has recently been introduced \cite{shao2025holitom,ju2026forestprune,fan2026flashvid,li2026aot,du2026unicomp}.

In this paper, we study token compression for efficient video-MLLMs from a new perspective, termed \emph{Visual Token Coding} (\textbf{VTC}).
Concretely, the principle of \vtc{} is inspired by the traditional video coding, such as HEVC \cite{sullivan2012hevc}.
Instead of transmitting all RGB frames, video coding organizes them
into \emph{Intra-coded} and \emph{Predictive-coded frames}, denoted as \textbf{I} and \textbf{P} frames, respectively.
As illustrated in Fig. \ref{fig:motivation}, an I-frame retains complete visual information and
is then used as a reference to structurally compress P frames through \emph{motion-compensated prediction}
\cite{sullivan2012hevc}.
During decoding, P frames can be restored through motion compensation and residual addition
using \emph{motion vectors} and \emph{prediction residuals}.
In this case, video coding can greatly reduce video size while minimizing information distortion.

\begin{figure}[t]
    \centering
    \includegraphics[width=\columnwidth]{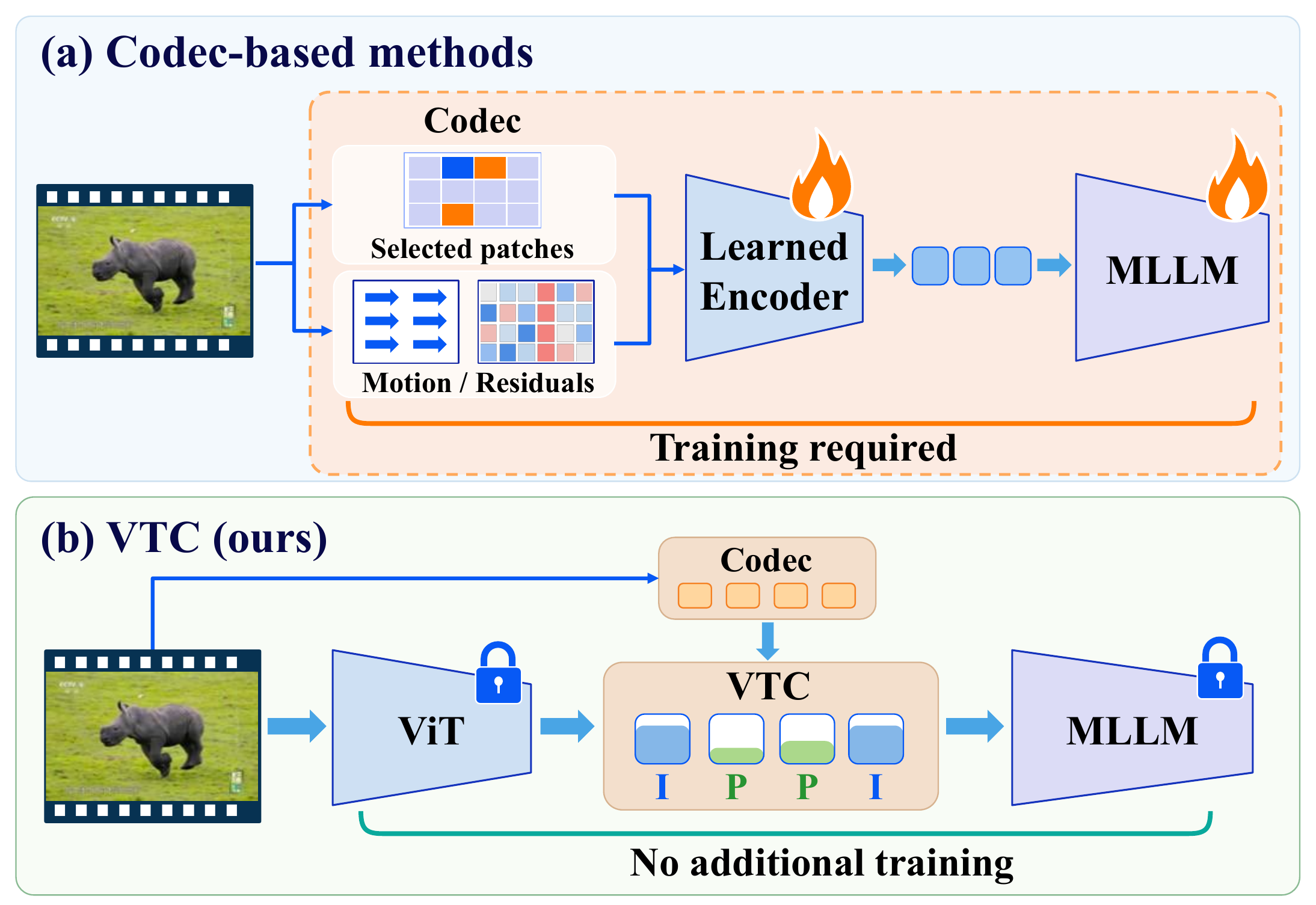}
    \caption{Comparison between recent \emph{codec}-based methods and our \vtc{}. (a) To borrow the merit of video coding, recent works compress the image patches or encode video motion and residuals based on the codec information, but both the vision encoder and the MLLM need to be trained to accommodate these codec inputs. (b) \wordflag{Our VTC directly performs structured compression at the token level}, yielding a \wordflag{training-free adaptation} for existing MLLMs.}
    \label{fig:codec-vtc-comparison}
    \vspace{-3pt}
\end{figure}

Given the similar objective of information compression, we think that the structured compression principle of video coding is applicable to video MLLMs, but a direct application is intractable.
First, while video data for benchmarking use video encoding \cite{zhou2025mlvu,wu2024longvideobench,wang2025lvbench,fu2025videomme}, video-MLLMs still need to input the reconstructed P frames, making it no different from directly using RGB video frames.
Second, the compressed structural information of video coding records dynamic changes between video frames, but this information does not provide suitable semantic information for the direct use of MLLMs.
In this case, attempts use selected patches or learned motion and residual features as video inputs \citep{tang2026ovencoder,an2026llavaov2,sarkar2026cope,yashima2026remora,hou2026adacodec}.
However, these \emph{codec}-based methods require additional training of both the vision encoder and the MLLM, limiting their direct applicability to existing MLLMs such as those in the Qwen-VL series ~\cite{bai2025qwen25vl,bai2025qwen3vl}.

To this end, we aim to apply the principle of video coding to token compression and achieve a general \emph{visual token coding} (\vtc{}) paradigm for existing MLLMs.
Specifically, we first build a base \vtc{} for token pruning based on the structure compression theory of video coding{.} \vtc{} also adopts the I/P frame partition and uses the token-wise residuals to represent the redundancy of P-frame tokens, as shown in Fig.~\ref{fig:motivation}. 
Given this base paradigm, we notice that it still has ample room to optimize, such as the proper budget allocation of P-frames' tokens and the redundancy of I-frames' ones. In this case, we further extend \vtc{} with a set of dynamic budgeting designs, termed \textbf{VTC\textsubscript{dy}}, achieving more cost-effective token coding for different videos.

Compared with traditional video coding, our \vtc{} approaches perform token-level compression and can be directly applied to existing MLLMs. Different from recent \emph{codec}-based methods \cite{tang2026ovencoder,sarkar2026cope,hou2026adacodec}, \vtc{} performs coding directly on visual tokens output by the visual encoders of MLLMs, and uses \emph{codec} information to allocate token budgets, as illustrated in Fig.~\ref{fig:motivation}. In this way, \vtc{} requires no additional training of the visual encoder or MLLM, yielding better generalization across MLLMs.
Moreover, \vtc{} also has desirable properties, such as \emph{query independence}~\cite{hyun2025sttm}, \emph{temporal modeling}~\cite{shao2025holitom}, and \emph{spatial integrity}~\cite{lin2026vcast}.

To validate the proposed \vtc{} paradigms, we apply them to three video MLLMs, namely \emph{Qwen3-VL-8B} \cite{bai2025qwen3vl}, \emph{LLaVA-OneVision} \cite{li2024llavaonevision} and \emph{LLaVA-OneVision-2} \cite{an2026llavaov2}, and also compare them with SOTA token compression methods \cite{chen2024fastv,yang2025visionzip,fan2026flashvid,shao2025holitom,fu2025framefusion,liu2025vidcom2,yuan2025unicomp} on multiple video benchmarks \cite{wu2024longvideobench,fu2025videomme,zhou2025mlvu,wang2025lvbench}.
The experimental results not only confirm the effectiveness of the \vtc{} paradigms for existing MLLMs, \emph{e.g.}, \vtc{} and \vtcdy{} retain 100.5\% and 102.1\% of the LLaVA-OneVision performance with a 50\% token retention ratio, but also 
show that \vtcdy{} achieves better performance and efficiency than the compared methods, \emph{e.g.}, 85.0\% lower pruning latency and a 1.8\% relative gain in average performance retention over the SOTA FlashVID\wordflag{~}\cite{fan2026flashvid} on Qwen3-VL.

Overall, our contributions are twofold.
\begin{itemize}
    \item We investigate the new paradigm of \emph{visual token coding} for video-MLLMs based on the principle of video coding, showing the feasibility of the structure compression theory for token pruning.

    \item Extensive experiments not only confirm the effectiveness of the proposed \vtc{} paradigms for video MLLMs but also demonstrate competitive or \wordflag{better performance than that of} existing methods with superior efficiency.
\end{itemize}


%% file: sec/2_related_work.tex
\section{Related Work}
\begingroup
Visual token compression reduces MLLM inference cost through pruning~\citep{zhang2025sparsevlm,xing2025pyramiddrop}, merging~\citep{bolya2023tome,shang2025prumerge}, or both~\citep{dhouib2025pact,yang2025libramerging,zhang2026hybrid}. FastV prunes tokens using attention scores~\citep{chen2024fastv}, while DivPrune uses token diversity~\citep{alvar2025divprune}. ToMe and LLaVA-PruMerge merge features~\citep{bolya2023tome,shang2025prumerge}, and VisionZip retains attention-selected tokens while merging the rest into context tokens~\citep{yang2025visionzip}. Our \vtc{} instead uses feature prediction residuals to retain information relative to preceding frames.

Video token compression addresses both intra-frame and temporal redundancy~\citep{tao2025dycoke,huang2025prunevid,liu2025vidcom2}. FrameFusion merges corresponding tokens across adjacent frames before pruning~\citep{fu2025framefusion}. HoliTom merges tokens within temporal segments~\citep{shao2025holitom}, while FlashVID selects tokens along a spatiotemporal tree and merges redundant features~\citep{fan2026flashvid}. UniComp combines frame merging, budget allocation by informational uniqueness, and within-frame compression~\citep{yuan2025unicomp}.

Codec-based methods use compressed-video information to improve video MLLM efficiency. OneVision-Encoder and LLaVA-OneVision-2 encode patches selected using codec information~\citep{tang2026ovencoder,an2026llavaov2}. CoPE-VideoLM encodes bitstream motion vectors and residuals as visual tokens~\citep{sarkar2026cope}. AdaCodec represents motion and prediction residuals as P-tokens~\citep{hou2026adacodec}, and ReMoRa refines motion representations from codec motion vectors~\citep{yashima2026remora}. These methods train visual encoders or codec-domain modules, fine-tune MLLMs, or both. MeToM uses codec residuals for pre-ViT patch merging and GoP sizes for budgeting~\citep{wu2026metom}. In contrast, \vtc{} applies I/P-frame coding directly to ViT-extracted tokens without additional training, with codec priors guiding token budgets in its dynamic variant.
\par
\endgroup

%% file: sec/3_preliminary.tex
\section{Preliminary}
\emph{Video coding}~\citep{sullivan2012hevc} compresses videos by exploiting spatial
and temporal redundancy across video frames, \emph{e.g.}, HEVC~\citep{sullivan2012hevc}.
Instead of encoding each frame independently, video coding organizes reference
information according to video continuity and uses \emph{predictive coding},
\emph{residual representation}, and \emph{resource allocation} to preserve critical
visual information under a limited coding budget.

Specifically, given a current video frame $I_t \in \mathrm{R}^{H\times W\times3}$, where $H\times W$ denotes the image resolution, video coding methods such as HEVC first predict the current frame from a historical reference frame $I_{\mathrm{ref}}$:
{\small
\begin{equation}
\label{eq:preliminary-prediction}
\widehat{I}_t = f(I_{\mathrm{ref}}),
\end{equation}%
}%
where $f(\cdot)$ denotes the inter-frame prediction process.
The difference between the current and predicted frames is then represented
as a residual $\Delta_t^{\mathrm{pix}}$:
{\small
\begin{equation}
\label{eq:preliminary-residual}
\Delta_t^{\mathrm{pix}} = I_t - \widehat{I}_t.
\end{equation}%
}%

Because the residual describes only the new changes relative to the reference
content, encoding this residual instead of the full frame can substantially reduce
temporal redundancy between consecutive frames.

The residual magnitude $\kappa_t^{\mathrm{pix}}$ then provides a simple measure of the change at the current
position:
{\small
\begin{equation}
\label{eq:preliminary-complexity}
\kappa_t^{\mathrm{pix}} = \|\Delta_t^{\mathrm{pix}}\|,
\end{equation}%
}%
where $t$ denotes the $t$-th time step. A larger residual magnitude commonly indicates motions, a scene change, or a region with
complex texture, and then it requires a larger coding budget for this time step.

This coding structure yields a compact video bitstream. An uncompressed video
of $N$
RGB frames at resolution $H\times W$ with $b$ bits per
channel requires a bit budget of
{\small
\begin{equation}
\label{eq:raw-bit-budget}
B_{\mathrm{raw}}=N\times(H\times W\times3b).
\end{equation}%
}%
And its size after coding becomes
{\small
\begin{equation}
\label{eq:coded-bit-budget}
B_{\mathrm{coded}}=\sum_{t\in\mathcal{I}_{\mathrm{c}}}b^t_{I}
+\sum_{t\in\mathcal{P}_{\mathrm{c}}}
\left(b^t_{\mathrm{mv}}+b^t_{\mathrm{res}}\right).
\end{equation}%
}%
Here, $\mathcal{I}_{\mathrm{c}}$ and $\mathcal{P}_{\mathrm{c}}$ are the codec I/P-frame index sets with
$N=|\mathcal{I}_{\mathrm{c}}|+|\mathcal{P}_{\mathrm{c}}|$, $b^t_{I}$ is the
intra-coded I-frame size, and $b^t_{\mathrm{mv}}$ and
$b^t_{\mathrm{res}}$ are the bit lengths of P-frame motion and residual
information.
These bit lengths provide frame-level coding costs, while the underlying frame types, motion vectors, and coded residuals constitute \emph{codec information} \cite{sullivan2012hevc}.
For predictable content, P-frame terms are much smaller
than the corresponding raw RGB data, yielding
$B_{\mathrm{coded}}\ll B_{\mathrm{raw}}$. 

This video coding principle suggests that efficient compression should jointly
consider reference information, newly introduced content, and the budget
allocation. However, motion vectors and pixel residuals do not match existing encoders' RGB inputs. Recent codec-based works  \citep{sarkar2026cope,an2026llavaov2} exploit this principle mainly based on the dedicated structural modifications and tuning of visual encoders, which limits their direct adaptation to existing MLLMs.

To bridge this gap, we reformulate video token compression as a \emph{token coding}
process and transfer the corresponding coding principles to the visual token
space.


%% file: sec/4_method.tex
\input{fig/framework}

\section{Method}

Motivated by video coding, we propose a novel paradigm termed \emph{Visual Token Coding} (\vtc{})
for the effective token compression of video MLLMs, and also explore its dynamic budgeting designs for more cost-effective token coding.

\subsection{Visual Token Coding}

We first build a base \emph{visual token coding} (\vtc{}) paradigm based on the principle of video coding, as illustrated in Fig.~\ref{fig:framework}.
Different from existing \emph{importance-}based \cite{chen2024fastv,yang2025visionzip} or \emph{relevance-}based methods \cite{huang2025prunevid,liu2025kvtp}, \vtc{} focuses on whether visual tokens can be predicted from previous frames or contain new information.

In this context, \vtc{} constructs semantic I/P frames by transferring \emph{reference prediction} and \emph{residual coding} to the visual token space, where I frames preserve scene references and P frames represent changes relative to them.

Given $N$ sampled video frames, the visual encoder extracts token features $F_t\in\mathbb{R}^{H'\times W'\times d}$ for each frame $t$. Here, $H'\times W'$ denotes the token grid size and $d$ is the feature dimension.
For a retention ratio $r\in(0,1]$, the total token budget of \vtc{} is


{\small
\begin{equation}
\label{eq:total-token-budget}
B_{\mathrm{total}}=\operatorname{round}\left(r\sum_{t=1}^{N}(H'\times W')\right),
\end{equation}%
}%

To achieve the \wordflag{above budget}, \vtc{} \wordflag{first computes predictions and residuals} frame by frame. 
We initialize the reference with $F_1$, giving $\widehat{F}_2=F_1$, and the second frame's residual $\Delta_2^{\mathrm{feat}}$ is computed by

{\small
\begin{equation}
\label{eq:initial-reference-prediction}
\Delta_2^{\mathrm{feat}}=F_2-\widehat{F}_2\wordflag{.}
\end{equation}%
}%
\wordflag{For} each time step $t \ge 2$, we use the preceding unpruned features $F_{\mathrm{ref}}=F_{t-1}$ as the reference. Following Eqs.~\ref{eq:preliminary-prediction}--\ref{eq:preliminary-residual}, prediction and residual computation in feature space are computed by
{\small
\begin{equation}
\label{eq:semantic-residual}
\Delta_t^{\mathrm{feat}}=F_t-\widehat{F}_t,
\quad \text{where} \quad
\widehat{F}_t=f(F_{\mathrm{ref}}).
\end{equation}%
}%

Here, $f$ is the identity map. We then measure the residual energy of each token $i$ by
{\small
\begin{equation}
\label{eq:semantic-residual-energy}
E_{t,i}=\|\Delta_t^{\mathrm{feat}}(i)\|_2,
\end{equation}%
}%
where $\|\cdot\|_2$ is the Euclidean norm.
Then, tokens with higher energies receive priority during selection.

From Eq.~\ref{eq:semantic-residual-energy}, a sharp increase in residual energy suggests that the current frame is poorly predicted, so it should be classified as an I frame.

In practice, we compare against recent residuals to account for the different levels of motion across videos.
In \vtc{}, the first frame is defined as a semantic I frame, and the later ones are classified using current and historical mean residuals, denoted as $E_t$ and $E_x$:
{\small
\begin{equation}
\label{eq:mean-residual-history}
E_t=\frac{1}{H'\times W'}\sum_{i=1}^{H'\times W'}E_{t,i},
\qquad
E_x=\frac{1}{|\mathcal{H}_t|}\sum_{j\in\mathcal{H}_t}E_j.
\end{equation}%
}%

Here, $E_j$ is the mean residual of a recent frame $j$, and $\mathcal{H}_t$ indexes the frames used to compute the historical baseline $E_x$.
Once two historical residuals are available, a frame is classified as I when $E_t>\tau E_x$, where $\tau>0$ is the change threshold. Other frames are classified as P.

In a streaming setting, each incoming frame is compared with the preceding frame's complete features using only previously observed information. The residual history will be reset after each I-frame decision.

\paragraph{Fixed Token Budgeting and Selection.}
Let $\mathcal{I}$ and $\mathcal{P}$ be the semantic I/P-frame index sets.
The base \vtc{} retains all I-frame tokens and divides the remaining budget evenly among P frames within their capacities.
Each P frame retains its Top-$K$ highest-residual tokens by
{\small
\begin{equation}
\label{eq:basic-budget-accounting}
\sum_{t\in\mathcal{I}}(H'\times W')+\sum_{t\in\mathcal{P}}K_t^P
=B_{\mathrm{total}}.
\end{equation}%
}%
Lastly, the retained tokens are input to the MLLM.

\subsection{VTC with Dynamic Budgeting}

The base \vtc{} establishes an effective token coding paradigm for video-MLLMs, but it still leaves ample room for improvement. For instance, the budgets of P-frames are fixed and inflexible, and \wordflag{retaining all I-frame tokens can preserve redundant information}. In this case, we explore more dynamic budgeting designs for \vtc{}, termed \vtcdy{}.


\paragraph{Codec-Guided Dynamic Budgeting.} One intuitive solution is to use \emph{codec bitcost} to better allocate token budgets, which is similar to classical video coding like HEVC~\cite{sullivan2012hevc} and is also applied by recent codec-based vision encoders~\cite{an2026llavaov2,tang2026ovencoder,sarkar2026cope,hou2026adacodec}.

As discussed above, this codec information can be used to indicate the content changes of P-frames. However, the default video codecs are obtained by comparing adjacent frames in videos with high frame rates, \emph{e.g.}, 30 FPS in HEVC~\cite{sullivan2012hevc}. In contrast, existing video-MLLMs often sparsely sample video frames for efficient inference, \emph{e.g.}, 1 FPS by Qwen3-VL~\cite{bai2025qwen3vl}. In this case, directly using the default bitcost will ignore the actual content changes between the sampled frames by video-MLLMs, potentially underestimating the required token budget.

To this end, we first align the codec information to the actual sampled video frames of video-MLLMs. Given two adjacent frames $V_{i-\delta}$ and $V_i$, where $\delta$ denotes the time interval, we collect the default codec bitcost of this interval, denoted as $b_\delta\in\mathbb R^{1\times\delta}$.

According to~\cite{tsang2012flash}, the default codec bitcost often involves large outliers due to the issues like abrupt illumination changes. Thus, we first apply a \emph{logarithmic transform} \cite{box1964transformations} to normalize these values:
{
\begin{equation}
\label{eq:interval-bitcost-log-normalization}
b_\delta^{\log}=\log\left(1+\frac{b_\delta}{H\times W}\right).
\end{equation}%
}%

Next, we compute the median $u$ and median absolute deviation $\sigma$~\cite{rousseeuw1993mad} from the log bitcosts of video frames. We use Softplus~\cite{dugas2000softplus} to map each element of $b_\delta^{\log}$ to a positive value, ensuring non-negative weights in budget allocation:
{\small
\begin{equation}
\label{eq:per-frame-normalized-bitcost}
b_\delta^{\mathrm{rel}}=
\operatorname{Softplus}\!\left(
\frac{b_\delta^{\log}-u}{\sigma+\epsilon}
\right)+\epsilon.
\end{equation}%
}%

\wordflag{Here, $u$ and $\sigma$ denote} the median and the median absolute deviation of the video's \wordflag{log-transformed bitcosts}. Then we can get the allocation weights $w_i$ by
{\small
\begin{equation}
\label{eq:temporal-complexity-score}
\begin{gathered}
w_i = \frac{b_i}{\sum_{j\in\mathcal P}b_j},\\
\text{where}\quad
b_i=\alpha\operatorname{Mean}(b_i^{\mathrm{rel}})
+(1-\alpha)Q_{0.9}(b_i^{\mathrm{rel}}).
\end{gathered}
\end{equation}%
}%
The mean reflects the overall coding cost within the interval, while the 90th percentile (P90)~\cite{hyndman1996samplequantiles} reflects higher local coding costs. Here, $w_i$ determines the share of the P-frame budget that $V_i$ receives beyond its minimum allocation.


\paragraph{Budgeted P-Frame Pruning.}
Using the allocation weights $w_t$ in Eq.~\ref{eq:temporal-complexity-score}, we determine the relative share of the remaining budget assigned to each semantic P frame. We first reserve a fixed budget $K_t^I=\operatorname{round}(\rho H'\times W')$ for each semantic I frame, where $\rho$ is the I-frame retention ratio. The remaining budget $B^P=B_{\mathrm{total}}-\sum_{t\in\mathcal I}K_t^I$ is allocated to semantic P frames. Each P frame is guaranteed a minimum allocation of $K_{\min}^P$ tokens, and the remaining budget is distributed accordingly:

{
\begin{equation}
\label{eq:dynamic-p-frame-budget}
K_t^P=K_{\min}^P+w_t\left(B^P-|\mathcal P|K_{\min}^P\right).
\end{equation}%
}%
\paragraph{Budgeted I-Frame Pruning.}
A semantic I frame serves as a scene reference for subsequent frames. It needs to retain a representative within-frame structure, but \wordflag{retaining all I-frame tokens can introduce redundancy}. Given the I-frame's budget $K_t^I$, we still follow the structure compression principle to measure the tokens' within-frame information, which can be defined by
{\small
\begin{equation}
\label{eq:intra-frame-difference-score}
\begin{aligned}
C_{t,i}=\|f_t^i-\bar{f}_t\|_2,
\end{aligned}
\end{equation}%
}%
where $f_t^i$ denotes the feature of token $i$ in the $t$-th frame, and $\bar{f}_t$ is the mean token feature of that frame. $C_{t,i}$ {measures the token's deviation from the frame mean to preserve within-frame structure.
We then combine the within-frame deviation and inter-frame residual to compute the importance of each token in a semantic I frame:}
{\small
\begin{equation}
\label{eq:semantic-i-frame-score}
S_{t,i}=\begin{cases}
\dfrac{E_{t,i}+C_{t,i}}{2}, & t>1,\\[4pt]
C_{1,i}, & t=1.
\end{cases}
\end{equation}%
}%
{Then, we use $S_{t,i}$ to retain $K_t^I$ tokens through the redundancy suppression described below.}

\paragraph{Frame Redundancy Suppression.}


{Global Top-K ranking may select multiple highly similar tokens from the same local region, leading to redundant representations within a frame. To mitigate this issue, we divide each frame into $M=G_hG_w$ regions and reserve a fraction $\rho_s$ of its token budget $K_t$ for region-wise selection:}

{\small
\begin{equation}
\label{eq:spatial-coverage-budget}
K_t^{\mathrm{cov}}=
\min\left(K_t,\max\left(M,\lfloor\rho_sK_t\rfloor\right)\right).
\end{equation}%
}%


For both I and P frames, we use $S_{t,i}$ and $E_{t,i}$ as token importance scores, respectively. We first allocate $K_t^{\mathrm{cov}}$ across spatial regions and select the highest-scoring tokens within each region to ensure spatial coverage. The remaining budget is filled by globally selecting the highest-scoring unselected tokens, preserving additional informative content.

%% file: fig/framework.tex
\begin{figure*}[t]
    \centering
    \includegraphics[width=0.98\textwidth,keepaspectratio]{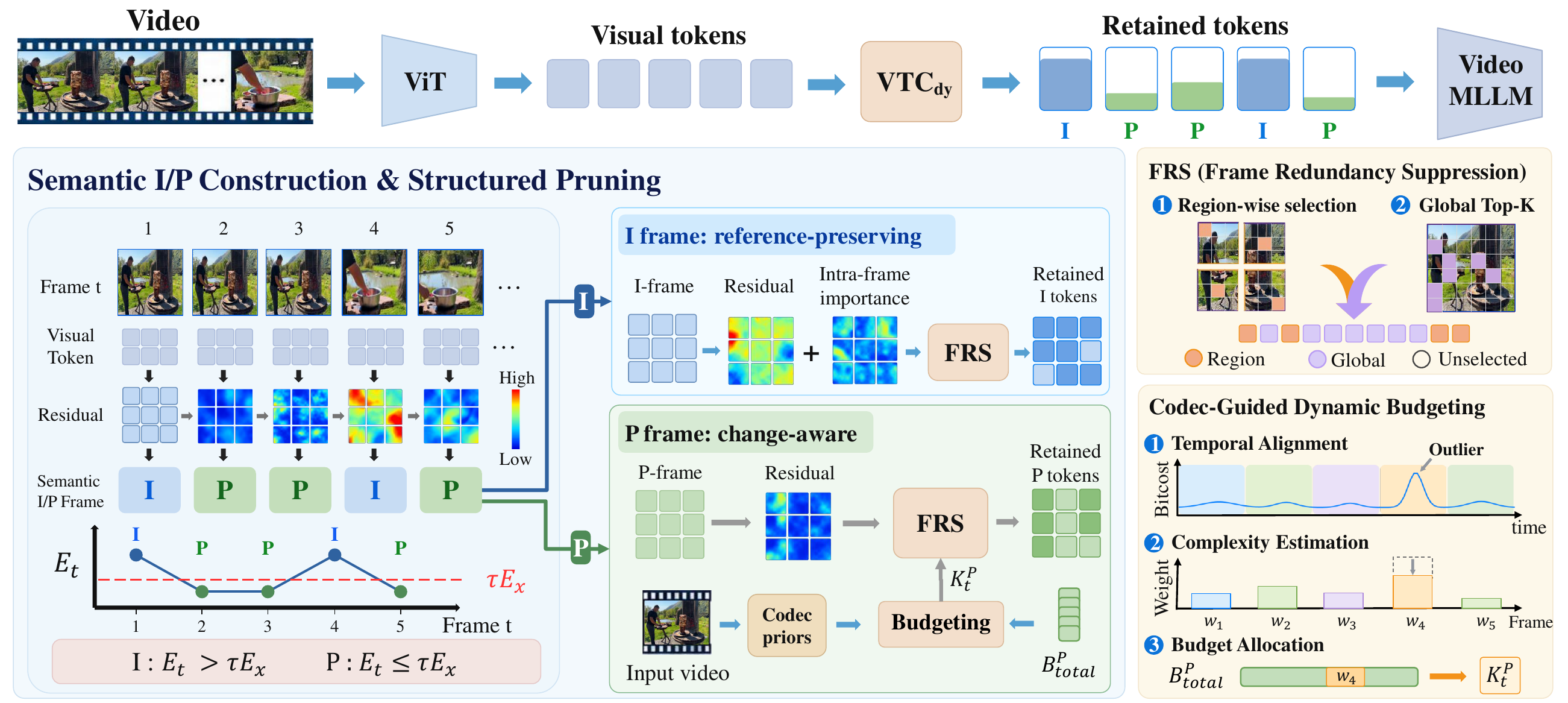}
    \caption{Overview of the proposed \emph{Visual Token Coding} (VTC) methods. Following the principle of video coding \cite{sullivan2012hevc}, the base VTC (the blue part) also identifies the I and P frames in videos, where the former preserve scene references and the latter {describe} the changes. The difference is that VTC performs structured compression on the visual token level, directly applicable to existing MLLMs. To achieve more cost-effective token coding, we further extend VTC with dynamic budgeting designs. We first align the default \emph{codec bitcosts} computed via a high frame rate (e.g., 30 FPS) to the sparsely sampled frames in MLLMs (e.g., 0.5--1 FPS), based on which the dynamic budgets are computed for each P frame (right bottom). Besides, I frames are further compressed following the same structured coding principle (middle). Lastly, \emph{Frame Redundancy Suppression} (FRS) is used for the final token selection.}
    \label{fig:framework}
\end{figure*}

%% file: sec/5_experiments.tex
\input{tabs/main_res}
\input{tabs/sota_qwen}

\section{Experiments}

\subsection{Experimental Settings}
 
\paragraph{Implementation Details.}
\label{sec:implementation}

We evaluate \vtc{} and \vtcdy{} using three MLLMs, namely Qwen3-VL-8B-Instruct \citep{bai2025qwen3vl}, LLaVA-OneVision-7B \citep{li2024llavaonevision}, and LLaVA-OneVision-2-8B \citep{an2026llavaov2} at 50\% and 25\% retention.
For Qwen3-VL-8B-Instruct, we use 64 input frames with the default \texttt{max\_pixels}. For LLaVA-OneVision-7B, we use 32 input frames with 196 tokens per frame; LLaVA-OneVision-2-8B uses 64 input frames and \texttt{max\_pixels}$=200\mathrm{K}$.
{\vtcdy{} takes Groups of Pictures as input, each containing one full-resolution image and four half-resolution images. All methods are matched in ViT computational cost and the total number of tokens fed into the MLLM.}
Most experiments use $\tau=2$, $\alpha=0.5$, $\rho=0.9$, equal weights for the two I-frame importance terms, and 16 FRS regions with $\rho_s=0.2$. A few benchmarks use minor hyperparameter adjustments.
\input{fig/performance_time}

\paragraph{Benchmarks and Metrics. }
We evaluate \vtcdy{} on four long-video understanding benchmarks, namely MLVU
\citep{zhou2025mlvu}, LongVideoBench \citep{wu2024longvideobench}, LVBench
\citep{wang2025lvbench}, and Video-MME \citep{fu2025videomme}.
They cover diverse genres and durations, from short clips to videos lasting up
to two hours. We report accuracy for all multiple-choice VideoQA tasks.


\ifdefined\isarxivversion
\setcounter{dbltopnumber}{1}
\fi
\begin{figure*}[t]
    \centering
    \ifdefined\isarxivversion
    \setlength{\abovecaptionskip}{6pt}
    \includegraphics[width=\textwidth,keepaspectratio]{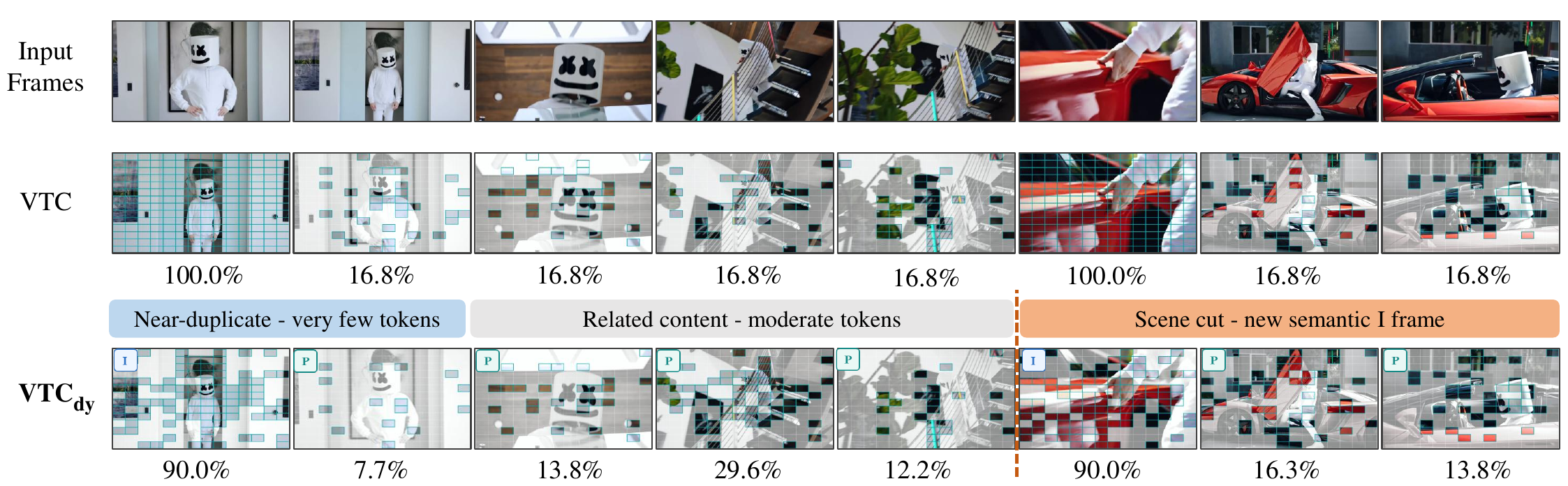}
    \else
    \includegraphics[width=\textwidth,keepaspectratio]{figs/case_study_v2.pdf}
    \fi
    \caption{Visualization of our \vtc{} methods on Qwen3-VL using 64 video frames. The base \vtc{} keeps {all tokens} of I frames and applies a fixed pruning ratio for all P frames. Although superior in coding redundant tokens, its static coding strategy {is} prone to insufficient budget allocations for informative P-frames. With more dynamic budgeting designs, \vtcdy{} can {offer} better token budgets for both I and P frames, achieving more cost-effective token coding for existing MLLMs.}
    \label{fig:case_study}
\end{figure*}

\subsection{Quantitative Analysis}
\paragraph{Effects of VTC on Different MLLMs.}
Tab.~\ref{tab:main_comparison} shows that the base \vtc{} provides effective token compression across all three MLLMs. At 50\% retention, it maintains near-full-model performance across all four benchmarks. Notably, on LLaVA-OneVision-7B, it preserves or slightly improves performance on every benchmark. However, under the more aggressive 25\% budget, the performance degradation becomes more pronounced on Qwen3-VL-8B and LLaVA-OneVision-2-8B, with average retention falling below 97\%. This suggests that retaining all I-frame tokens and uniformly allocating the P-frame budget are insufficient to consistently preserve video information under tight budgets. In contrast, \vtcdy{} combines codec-guided dynamic P-frame allocation, lightweight I-frame compression, and intra-frame redundancy suppression, improving performance retention across all three MLLMs at both budgets. Particularly at 25\% retention, it maintains near-full-model performance across all three backbones. These results support our motivation and demonstrate the benefits of adapting frame-level budgets to temporal complexity under constrained token budgets.

\paragraph{Comparison with SOTA Compression Methods.}
Tab.~\ref{tab:sota_qwen} compares \vtc{} with representative visual token compression methods under matched pre-pruning visual encoding costs. Despite its simple design, the base \vtc{} achieves competitive performance against recent approaches. For example, on Qwen3-VL-8B at 25\% retention, it achieves 95.9\% average performance retention, outperforming FrameFusion, HoliTom, and UniComp. This result further demonstrates the effectiveness of the proposed token coding paradigm, even without additional dynamic allocation designs. Building on this foundation, \vtcdy{} achieves further improvements on both MLLMs and obtains the best results on most benchmarks. Notably, at 25\% retention, it improves the average performance retention over the recent FlashVID by 1.8\% on both Qwen3-VL-8B 
and LLaVA-OneVision-7B. 
These results \wordflag{demonstrate} the effectiveness of our structured token coding paradigm and validate codec-guided dynamic budget allocation under aggressive compression.

\paragraph{Analysis of Efficiency.} Fig. ~\ref{fig:performance_time} compares average performance retention and inverse latency on Qwen3-VL at a 25\% token retention ratio.\footnote{Tested on one RTX 3090 with 64 frames as input.} Latency includes both token pruning and LLM prefill. Among the compared compression methods, \vtcdy{} achieves the highest performance retention of 98.3\% and the lowest latency. Compared with the SOTA method FlashVID \cite{fan2026flashvid}, \vtcdy{} improves retention by 1.8\% while running 1.34x faster under this setting. The reason is that FlashVID requires the pairwise similarity computation with a complexity of $\mathcal{O}(N^2)$, while our structured compression is only $\mathcal{O}(N)$. \wordflag{These results support the efficiency of VTC methods while maintaining high performance.}

\ifdefined\isarxivversion\else
\input{tabs/ab_all}
\fi

\ifdefined\isarxivversion
\flushbottom
\fi
\paragraph{Ablation Study.}
We conduct ablation studies to examine the individual contributions of \emph{P-frame budgeting}, \emph{I-frame budgeting}, and \emph{FRS} in Tab.~\ref{tab:main_ablation}.
\ifdefined\isarxivversion
\input{tabs/ab_all}
\fi
Adding P-frame budgeting improves performance, particularly on MLVU Multi-Detail, demonstrating the benefit of allocating tokens according to temporal complexity rather than treating all P-frames equally. Further introducing I-frame budgeting brings consistent improvements across most metrics. By compressing I-frames at a fixed ratio, it avoids excessive token consumption by reference frames and allows more resources to be allocated to P-frames. Finally, FRS further improves MLVU Single-Detail, Multi-Detail, and LVBench by reducing redundant token selection within individual frames. Although Holistic performance slightly decreases, the overall MLVU average continues to improve. Overall, the full method achieves the best results on most metrics, demonstrating the complementary roles of temporal budget allocation and intra-frame redundancy suppression in improving token utilization of our VTC paradigms.
\paragraph{Dynamic Token Allocation Strategy.}
The last block of Tab.~\ref{tab:main_ablation} investigates different temporal complexity statistics used to compute the allocation weights in Eq.~\ref{eq:temporal-complexity-score}. \emph{Uniform} assigns an equal budget to all P-frames, ignoring variations in content complexity across temporal intervals. Among the single-statistic strategies, \emph{Bitcost mean} captures overall coding complexity and performs best on MLVU Holistic, while \emph{Bitcost P90} emphasizes local complexity peaks and shows advantages on MLVU Multi-Detail and LVBench. These results suggest that the two statistics capture complementary aspects of temporal complexity. By combining them, our method achieves the best results across most benchmarks. This demonstrates that jointly considering overall coding complexity and local complexity variations leads to more balanced token allocation.

\subsection{Qualitative Analysis}

Fig.~\ref{fig:case_study} visualizes how \vtcdy{} redistributes tokens along a video timeline. The baseline \vtc{} uniformly assigns 16.8\% of the token budget to every P-frame, irrespective of temporal novelty. In contrast, \vtcdy{} allocates only 7.7\% to a near-duplicate second frame, allocates moderate budgets to related content, and increases the budget to 29.6\% for a frame with substantial content changes. A large residual at a scene transition triggers the classification of the position as a new semantic I frame. This distinction further guides the allocation of frame-specific token budgets. Furthermore, the token masks reveal that \vtcdy{} preserves richer visual evidence around content changes while  suppressing redundant information in stable intervals, confirming its ability to adaptively allocate resources based on semantic importance.

%% file: tabs/main_res.tex
\begin{table*}[t]
\centering
\caption{
Results on three video MLLMs. Retain is relative to the uncompressed backbone;
Avg. Retain averages four benchmarks. Bold marks the better result at each
retention rate. Comparisons match pre-pruning visual encoding cost.
}
\label{tab:main_comparison}

\small
\setlength{\tabcolsep}{7.3pt}
\renewcommand{\arraystretch}{1.02}

\begin{tabular}{
lc
cc
cc
cc
cc
c}
\toprule
\multirow{2}{*}{Method}
& \multirow{2}{*}{\shortstack{Retention\\Rate}}
& \multicolumn{2}{c}{LVB}
& \multicolumn{2}{c}{Video-MME}
& \multicolumn{2}{c}{MLVU}
& \multicolumn{2}{c}{LVBench}
& \multirow{2}{*}{\shortstack{Avg.\\Retain}} \\
\cmidrule(lr){3-4}
\cmidrule(lr){5-6}
\cmidrule(lr){7-8}
\cmidrule(lr){9-10}
&
& Acc. & Retain
& Acc. & Retain
& Acc. & Retain
& Acc. & Retain
& \\
\midrule

\rowcolor{black!8}
\textit{Qwen3-VL-8B}
& 1.00
& 62.2 & 100.0\%
& 67.2 & 100.0\%
& 70.3 & 100.0\%
& 43.6 & 100.0\%
& 100.0\% \\

\vtc{}
& 0.50
& 61.9 & 99.5\%
& 66.7 & 99.3\%
& 69.3 & 98.6\%
& 42.0 & 96.3\%
& 98.4\% \\

\vtcdy{}
& 0.50
& \textbf{63.3} & \textbf{101.8\%}
& \textbf{66.8} & \textbf{99.4\%}
& \textbf{70.3} & \textbf{100.0\%}
& \textbf{43.6} & \textbf{100.0\%}
& \textbf{100.3\%} \\

\cmidrule(lr){1-11}

\vtc{}
& 0.25
& 61.3 & 98.6\%
& 65.3 & 97.2\%
& 67.3 & 95.7\%
& 40.1 & 92.0\%
& 95.9\% \\

\vtcdy{}
& 0.25
& \textbf{62.5} & \textbf{100.5\%}
& \textbf{66.2} & \textbf{98.5\%}
& \textbf{69.4} & \textbf{98.7\%}
& \textbf{41.6} & \textbf{95.4\%}
& \textbf{98.3\%} \\

\midrule
\midrule

\rowcolor{black!8}
\textit{LLaVA-OneVision-7B} & 1.00 & 56.2 & 100.0\% & 58.4 & 100.0\% & 63.2 & 100.0\% & 37.8 & 100.0\% & 100.0\% \\
\cmidrule(lr){1-11}
\vtc{} & 0.50 & 56.3 & 100.2\% & 59.3 & 101.5\% & 63.3 & 100.2\% & 37.8 & 100.0\% & 100.5\% \\
\vtcdy{} & 0.50 & \textbf{56.6} & \textbf{100.7\%} & \textbf{59.6} & \textbf{102.1\%} & \textbf{64.7} & \textbf{102.4\%} & \textbf{39.1} & \textbf{103.4\%} & \textbf{102.1\%} \\
\cmidrule(lr){1-11}
\vtc{} & 0.25 & 56.4 & 100.4\% & 58.7 & 100.5\% & 62.9 & 99.5\% & 37.3 & 98.7\% & 99.8\% \\
\vtcdy{} & 0.25 & \textbf{57.6} & \textbf{102.5\%} & \textbf{59.3} & \textbf{101.5\%} & \textbf{63.8} & \textbf{101.0\%} & \textbf{39.1} & \textbf{103.4\%} & \textbf{102.1\%} \\
\midrule
\midrule

\rowcolor{black!8}
\textit{LLaVA-OneVision-2-8B}
& 1.00
& 64.1 & 100.0\%
& 65.7 & 100.0\%
& 72.3 & 100.0\%
& 45.1 & 100.0\%
& 100.0\% \\

\vtc{}
& 0.50
& 63.8 & 99.5\%
& 65.4 & 99.5\%
& 71.4 & 98.8\%
& 43.5 & 96.5\%
& 98.6\% \\

\vtcdy{}
& 0.50
& \textbf{64.5} & \textbf{100.6\%}
& \textbf{66.2} & \textbf{100.8\%}
& \textbf{71.9} & \textbf{99.5\%}
& \textbf{45.2} & \textbf{100.2\%}
& \textbf{100.3\%} \\

\cmidrule(lr){1-11}

\vtc{}
& 0.25
& \textbf{62.4} & \textbf{97.4\%}
& 64.4 & 98.0\%
& 70.0 & 96.8\%
& 41.9 & 92.9\%
& 96.3\% \\

\vtcdy{}
& 0.25
& 61.6 & 96.1\%
& \textbf{64.5} & \textbf{98.2\%}
& \textbf{71.3} & \textbf{98.6\%}
& \textbf{44.4} & \textbf{98.5\%}
& \textbf{97.8\%} \\

\bottomrule
\end{tabular}

\end{table*}

%% file: tabs/sota_qwen.tex
\begin{table*}[t]
\centering
\caption{
Comparison with token compression methods on Qwen3-VL-8B and LLaVA-OneVision-7B. \emph{Retain} is relative
to the uncompressed backbone, and \emph{Avg. Retain} averages the results of four benchmarks. Bold
and underline mark the best and second-best compressed results. 
Comparisons match pre-pruning visual encoding cost. LVB denotes LongVideoBench.
}
\label{tab:sota_qwen}
\label{tab:sota_comparison}

\footnotesize
\setlength{\tabcolsep}{2.4pt}
\renewcommand{\arraystretch}{0.95}
\vspace{-5pt}

\begin{tabular*}{\textwidth}{@{\extracolsep{\fill}}
cl
cc
cc
cc
cc
c}
\toprule
\multirow{2}{*}{\shortstack{Retention\\Rate}}
& \multirow{2}{*}{Method}
& \multicolumn{2}{c}{LVB}
& \multicolumn{2}{c}{Video-MME}
& \multicolumn{2}{c}{MLVU}
& \multicolumn{2}{c}{LVBench}
& \multirow{2}{*}{\shortstack{Avg.\\Retain}} \\
\cmidrule(lr){3-4}
\cmidrule(lr){5-6}
\cmidrule(lr){7-8}
\cmidrule(lr){9-10}
&
& Acc. & Retain
& Acc. & Retain
& Acc. & Retain
& Acc. & Retain
& \\
\midrule

\multicolumn{11}{c}{\textbf{Qwen3-VL-8B}} \\
\midrule
1.00
& Full model
& 62.2 & 100.0\%
& 67.2 & 100.0\%
& 70.3 & 100.0\%
& 43.6 & 100.0\%
& 100.0\% \\

\midrule
\multirow{8}{*}{0.50} & VisionZip$_\text{CVPR25}$~\cite{yang2025visionzip} & 61.7 & 99.2\% & 66.7 & 99.3\% & 68.9 & 98.0\% & \underline{43.3} & \underline{99.2\%} & 98.9\% \\
  & FrameFusion$_\text{ICCV25}$~\cite{fu2025framefusion} & \underline{62.4} & \underline{100.3\%} & 66.2 & 98.5\% & 69.8 & 99.3\% & 41.8 & 95.9\% & 98.5\% \\
  & VidCom$^2$$_\text{EMNLP25}$~\cite{liu2025vidcom2} & 62.0 & 99.7\% & 66.8 & 99.4\% & 69.5 & 98.9\% & 43.0 & 98.6\% & \underline{99.1\%} \\
  & HoliTom$_\text{NeurIPS25}$~\cite{shao2025holitom} & 61.4 & 98.7\% & 66.6 & 99.1\% & 68.7 & 97.7\% & 42.2 & 96.8\% & 98.1\% \\
  & UniComp$_\text{CVPR26}$~\cite{yuan2025unicomp} & 61.5 & 98.9\% & \underline{66.9} & \underline{99.6\%} & \underline{69.9} & \underline{99.4\%} & 42.6 & 97.7\% & 98.9\% \\
  & FlashVID$_\text{ICLR26}$~\cite{fan2026flashvid} & 62.1 & 99.8\% & \textbf{67.0} & \textbf{99.7\%} & 68.0 & 96.8\% & 42.9 & 98.5\% & 98.7\% \\
  & \vtc{}(ours) & 61.9 & 99.5\% & 66.7 & 99.3\% & 69.3 & 98.6\% & 42.0 & 96.3\% & 98.4\% \\
  & \vtcdy{}(ours) & \textbf{63.3} & \textbf{101.8\%} & 66.8 & 99.4\% & \textbf{70.3} & \textbf{100.0\%} & \textbf{43.6} & \textbf{100.0\%} & \textbf{100.3\%} \\
\midrule
\multirow{8}{*}{0.25} & VisionZip$_\text{CVPR25}$~\cite{yang2025visionzip} & 59.9 & 96.3\% & 65.7 & 97.8\% & 67.3 & 95.8\% & 41.0 & 94.0\% & 96.0\% \\
  & FrameFusion$_\text{ICCV25}$~\cite{fu2025framefusion} & 61.2 & 98.4\% & 62.8 & 93.5\% & \underline{68.5} & \underline{97.4\%} & 40.5 & 92.9\% & 95.5\% \\
  & VidCom$^2$$_\text{EMNLP25}$~\cite{liu2025vidcom2} & 60.1 & 96.6\% & 65.3 & 97.2\% & 66.9 & 95.2\% & 41.0 & 94.0\% & 95.8\% \\
  & HoliTom$_\text{NeurIPS25}$~\cite{shao2025holitom} & 60.4 & 97.1\% & \underline{65.9} & \underline{98.1\%} & 66.1 & 94.0\% & 40.5 & 93.0\% & 95.6\% \\
  & UniComp$_\text{CVPR26}$~\cite{yuan2025unicomp} & 58.9 & 94.7\% & 65.6 & 97.6\% & 67.0 & 95.3\% & 41.2 & 94.5\% & 95.5\% \\
  & FlashVID$_\text{ICLR26}$~\cite{fan2026flashvid} & 60.4 & 97.0\% & 65.4 & 97.4\% & 66.7 & 94.9\% & \textbf{42.2} & \textbf{96.8\%} & \underline{96.5\%} \\
  & \vtc{}(ours) & \underline{61.3} & \underline{98.6\%} & 65.3 & 97.2\% & 67.3 & 95.7\% & 40.1 & 92.0\% & 95.9\% \\
  & \vtcdy{}(ours) & \textbf{62.5} & \textbf{100.5\%} & \textbf{66.2} & \textbf{98.5\%} & \textbf{69.4} & \textbf{98.7\%} & \underline{41.6} & \underline{95.4\%} & \textbf{98.3\%} \\
\midrule
\multicolumn{11}{c}{\textbf{LLaVA-OneVision-7B}} \\
\midrule
1.00 & Full model & 56.2 & 100.0\% & 58.4 & 100.0\% & 63.2 & 100.0\% & 37.8 & 100.0\% & 100.0\% \\
\midrule
\multirow{8}{*}{0.50} & VisionZip$_\text{CVPR25}$~\cite{yang2025visionzip} & 56.0 & 99.7\% & 59.0 & 101.0\% & 64.5 & 102.0\% & \underline{38.9} & \underline{102.9\%} & 101.4\% \\
  & FrameFusion$_\text{ICCV25}$~\cite{fu2025framefusion} & 56.6 & 100.7\% & 58.7 & 100.5\% & 64.0 & 101.3\% & 38.4 & 101.6\% & 101.0\% \\
  & VidCom$^2$$_\text{EMNLP25}$~\cite{liu2025vidcom2} & \textbf{56.8} & \textbf{101.1\%} & 58.4 & 100.0\% & 63.8 & 101.0\% & 38.6 & 102.1\% & 101.0\% \\
  & HoliTom$_\text{NeurIPS25}$~\cite{shao2025holitom} & 56.6 & 100.7\% & 58.7 & 100.5\% & \textbf{65.4} & \textbf{103.5\%} & 38.5 & 101.9\% & \underline{101.6\%} \\
  & UniComp$_\text{CVPR26}$~\cite{yuan2025unicomp} & \underline{56.7} & \underline{100.9\%} & 57.6 & 98.6\% & 64.2 & 101.6\% & 38.2 & 101.1\% & 100.5\% \\
  & FlashVID$_\text{ICLR26}$~\cite{fan2026flashvid} & 56.5 & 100.5\% & 58.4 & 100.0\% & 63.5 & 100.5\% & 38.2 & 101.1\% & 100.5\% \\
  & \vtc{}(ours) & 56.3 & 100.2\% & \underline{59.3} & \underline{101.5\%} & 63.3 & 100.2\% & 37.8 & 100.0\% & 100.5\% \\
  & \vtcdy{}(ours) & 56.6 & 100.7\% & \textbf{59.6} & \textbf{102.1\%} & \underline{64.7} & \underline{102.4\%} & \textbf{39.1} & \textbf{103.4\%} & \textbf{102.1\%} \\
\midrule
\multirow{8}{*}{0.25} & VisionZip$_\text{CVPR25}$~\cite{yang2025visionzip} & 55.7 & 99.2\% & 58.2 & 99.6\% & 64.2 & 101.5\% & 37.6 & 99.4\% & 99.9\% \\
  & FrameFusion$_\text{ICCV25}$~\cite{fu2025framefusion} & \underline{56.6} & \underline{100.7\%} & 58.7 & 100.5\% & 61.7 & 97.6\% & 37.8 & 100.0\% & 99.7\% \\
  & VidCom$^2$$_\text{EMNLP25}$~\cite{liu2025vidcom2} & 54.9 & 97.7\% & 58.6 & 100.3\% & 62.5 & 98.9\% & 36.8 & 97.4\% & 98.6\% \\
  & HoliTom$_\text{NeurIPS25}$~\cite{shao2025holitom} & 56.3 & 100.2\% & \underline{59.0} & \underline{101.0\%} & \underline{64.4} & \underline{101.9\%} & \underline{38.8} & \underline{102.7\%} & 101.4\% \\
  & UniComp$_\text{CVPR26}$~\cite{yuan2025unicomp} & \textbf{57.6} & \textbf{102.5\%} & 58.9 & 100.9\% & \textbf{65.0} & \textbf{102.9\%} & 37.7 & 99.7\% & \underline{101.5\%} \\
  & FlashVID$_\text{ICLR26}$~\cite{fan2026flashvid} & 56.2 & 100.0\% & \underline{59.0} & \underline{101.0\%} & 62.9 & 99.5\% & 38.1 & 100.8\% & 100.3\% \\
  & \vtc{}(ours) & 56.4 & 100.4\% & 58.7 & 100.5\% & 62.9 & 99.5\% & 37.3 & 98.7\% & 99.8\% \\
  & \vtcdy{}(ours) & \textbf{57.6} & \textbf{102.5\%} & \textbf{59.3} & \textbf{101.5\%} & 63.8 & 101.0\% & \textbf{39.1} & \textbf{103.4\%} & \textbf{102.1\%} \\
\bottomrule
\end{tabular*}

\end{table*}

%% file: fig/performance_time.tex
\begin{figure}[t]
    \centering
    \includegraphics[width=\columnwidth]{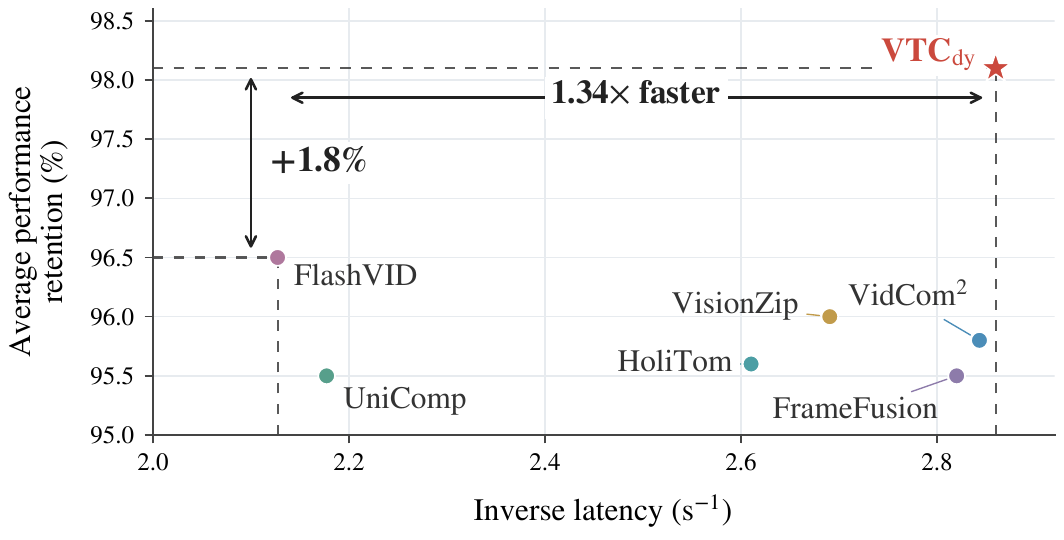}
    \vspace{-15pt}
    \caption{Performance–latency trade-off on Qwen3-VL-8B (64 frames) at 25\% retention on one RTX 3090. Latency includes pruning and MLLM prefilling, and the upper right is better.}
    \label{fig:performance_time}
    \ifdefined\isarxivversion
    \vspace{0pt}
    \else
    \vspace{-10pt}
    \fi
\end{figure}

%% file: tabs/ab_all.tex
\begin{table}[t]
\centering
\caption{Ablation on Qwen3-VL-8B under matched final-token and pre-pruning encoding budgets. $\dagger$ marks the default. All variants use the same input construction and total token budget. Superscripts show changes from the base row.}
\label{tab:main_ablation}
\label{tab:p_budget_signal}

\footnotesize
\setlength{\tabcolsep}{2pt}
\ifdefined\isarxivversion
\renewcommand{\arraystretch}{1.20}
\vspace{5pt}
\else
\renewcommand{\arraystretch}{0.92}
\vspace{-3pt}
\fi
\begin{tabular}{lccccc}
\toprule
\multirow{2}{*}{Choices}
& \multirow{2}{*}{LVBench}
& \multicolumn{4}{c}{MLVU} \\
\cmidrule(lr){3-6}
&
& S-Det.
& M-Det.
& Hol.
& M-avg \\
\midrule
\multicolumn{6}{c}{\textbf{Design Ablation}} \\
\midrule

Base
& 40.9
& 70.9
& 46.2
& 81.4
& 68.2 \\

+ P budgeting
& 41.1\textsuperscript{\tiny $\mathord{\uparrow}0.2$}
& 71.3\textsuperscript{\tiny $\mathord{\uparrow}0.4$}
& 50.5\textsuperscript{\tiny $\mathord{\uparrow}4.3$}
& 81.4\textsuperscript{\tiny $-0.0$}
& 68.9\textsuperscript{\tiny $\mathord{\uparrow}0.7$} \\

\hspace*{0.6em}+ I budgeting
& 41.1\textsuperscript{\tiny $\mathord{\uparrow}0.2$}
& 71.3\textsuperscript{\tiny $\mathord{\uparrow}0.4$}
& 50.9\textsuperscript{\tiny $\mathord{\uparrow}4.7$}
& \textbf{81.6}\textsuperscript{\tiny $\mathord{\uparrow}0.2$}
& 69.1\textsuperscript{\tiny $\mathord{\uparrow}0.9$} \\

\hspace*{1.2em}+ FRS
& \textbf{41.6}\textsuperscript{\tiny $\mathord{\uparrow}0.7$}
& \textbf{71.8}\textsuperscript{\tiny $\mathord{\uparrow}0.9$}
& \textbf{51.2}\textsuperscript{\tiny $\mathord{\uparrow}5.0$}
& 81.2\textsuperscript{\tiny $\mathord{\downarrow}0.2$}
& \textbf{69.4}\textsuperscript{\tiny $\mathord{\uparrow}1.2$} \\





\midrule
\multicolumn{6}{c}{\textbf{Dynamic Token Allocation Strategy}} \\
\midrule

Uniform
& 40.9
& 71.2
& 49.9
& 81.0
& 68.7 \\

Bitcost mean
& 41.3
& 71.0
& 49.9
& \textbf{82.1}
& 68.9 \\

Bitcost P90
& 41.4
& 70.9
& 50.1
& 81.4
& 68.8 \\


Bitcost mean + P90$^{\dagger}$
& \textbf{41.6}
& \textbf{71.8}
& \textbf{51.2}
& 81.2
& \textbf{69.4} \\





\bottomrule
\end{tabular}

\ifdefined\isarxivversion
\vspace{6pt}
\else
\vspace{-10pt}
\fi
\end{table}

%% file: sec/6_conclusion.tex
\section{Conclusion}



In this paper, we study the effective token compression of MLLMs from a new perspective of visual token coding (\vtc{}). 
Inspired by the classical principle of video coding, we first build the \vtc{} {paradigm that performs structured compression at the token level}, directly applicable to existing MLLMs. 
Afterwards, we also extend \vtc{} with more dynamic budgeting designs towards more cost-effective token coding, termed \vtcdy{}. 
Experiments support \vtc{} across three MLLMs and show a favorable accuracy--latency trade-off on Qwen3-VL, with latency measured over token pruning and MLLM prefill.


%% file: sec/llm_usage.tex
\section*{LLM Usage Statement}
We used OpenAI Codex at the Assist level to refine author-produced text, related work and citations, experimental design, execution and analysis, figures and visual materials, and research ideas, hypotheses, claims and conclusions. We used Codex at the Generate level for code and implementation, which the authors reviewed, edited or approved. No LLMs or AI tools were used for synthetic data, dataset creation or annotation. The authors take responsibility for all submitted content and results.